\pdfoutput=1

\documentclass[11pt]{article}

\usepackage[]{EMNLP2022}

\usepackage{times}
\usepackage{latexsym}

\usepackage[T1]{fontenc}

\usepackage[utf8]{inputenc}

\usepackage{microtype}

\usepackage{inconsolata}

\usepackage{amsmath}
\usepackage{booktabs}
\usepackage{graphicx}
\usepackage{amssymb}

\usepackage{tipa}
\usepackage{caption}
\usepackage{subcaption}

\usepackage{multirow}

\title{A Comprehensive Comparison of Neural Networks as Cognitive
Models of Inflection}

\author{Adam Wiemerslage \and Shiran Dudy \and Katharina Kann \\
University of Colorado Boulder \\
  \texttt{first.last@colorado.edu} \\}

\begin{document}
\maketitle
\begin{abstract}
Neural networks have long been at the center of a debate around the cognitive mechanism by which humans process inflectional morphology. This debate has gravitated into NLP by way of the question: Are neural networks a feasible account for human behavior in morphological inflection? 
We address that question by measuring the correlation between human judgments and neural network probabilities for unknown word inflections. We test a larger range of architectures than previously studied on two important tasks for the cognitive processing debate: English past tense, and German number inflection. We find evidence that the Transformer may be a better account of human behavior than LSTMs on these datasets, and that LSTM features known to increase inflection accuracy do not always result in more human-like behavior.
\end{abstract}

\section{Introduction: The Past Tense Debate}

Morphological inflection has historically been a proving ground for studying models of language acquisition. \citet{rumelhart1985learning} famously presented a neural network that they claimed could learn English past tense inflection. 
However, \citet{pinker1988language} proposed a dual-route theory for inflection, wherein regular verbs are inflected 
based on rules and irregular verbs are looked up in the lexicon. They highlighted several shortcomings of \citet{rumelhart1985learning} that they claimed any neural network would suffer from. 

This opened a line of work wherein cognitive theories of inflection are analyzed by implementing them as computational models and comparing their behavior to that of humans.
A famous study in the area of morphology is the \textit{wug test} \cite{berko1958child}, where human participants are prompted with a novel-to-them nonce word and asked to produce its plural form.
\begin{figure}[t]
  \centering
  \includegraphics[width=\columnwidth]{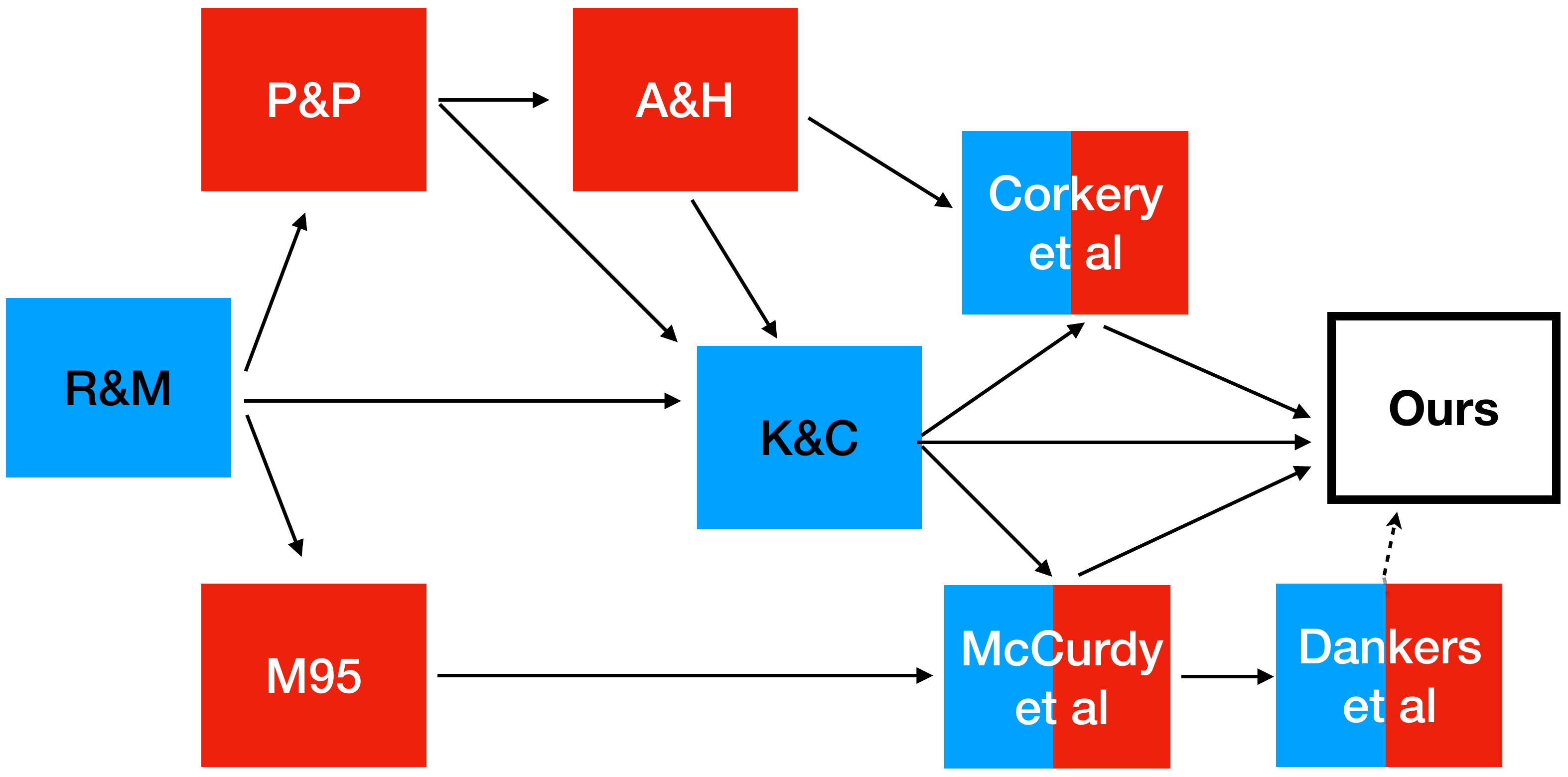}  
  \caption{Summary of the past tense debate as it pertains to this work, color coded by evidence for (blue) or against (red) neural networks as a cognitively plausible account for human behavior.}
  \label{fig:pst_tense_debate}
\end{figure}
Similarly, morphological inflection models are generally evaluated on words they have not seen during training.
However, since they are evaluated on actual words, it is impossible to meaningfully ask a native speaker, who knows the words' inflected forms, 
how likely different reasonable inflections for the words in a model's test data are.
Thus, in order to compare the behavior of humans and models on words unknown to both, prior work 
has created sets of made-up nonce words \cite{marcus1995german,albright2003rules}. 

\paragraph{English Past Tense}
English verbs inflect to express the past and present tense distinction. Most verbs inflect for past tense by applying the /-\textipa{d}/, /-\textipa{Id}/, or /-\textipa{t}/ suffix: allophones of the regular inflection class. Some verbs, however, express the past tense with a highly infrequent or completely unique inflection, forming the irregular inflection class. This distinction between regular and irregular inflection has motivated theories like the dual-route theory described above.

\citet{prasada1993generalisation} performed a wug test for English past tense inflection in order to compare the model from \citet{rumelhart1985learning} to humans with special attention to how models behave with respect to regular vs. irregular forms, finding that it could not account for human generalizations. \citet[A\&H]{albright2003rules} 
gathered production probabilities -- i.e., the normalized frequencies of the inflected forms produced by participants -- and ratings -- i.e., the average rating assigned to a given past tense form 
on a well-formedness scale. They then implemented two computational models: 
a rule-based and an analogy-based model 
and computed the correlation between the probabilities of past tense forms for nonce verbs under each model and according to humans. 
They found that the rule-based model more accurately accounts for nonce word inflection.

After several years of progress for neural networks, including 
state-of-the-art results on morphological inflection \cite{kann-schutze-2016-single,cotterell-etal-2016-sigmorphon}, this debate was revisited by \citet[K\&C]{kirov2018recurrent}, who examined modern neural networks. They trained a bidirectional LSTM \cite{hochreiter1997long} with attention \cite{bahdanau2015neural} on English past tense inflection and in experiments quantifying model accuracy on a held out set of real English verbs, they showed that it addresses many of the shortcomings pointed out by \citet{pinker1988language}. They concluded that the LSTM is, in fact, capable of modeling English past tense inflection.
They also applied the model to the wug experiment from A\&H
and found a 
positive correlation with human production probabilities that was slightly higher than the rule-based model from A\&H.

\citet[C\&al.]{corkery2019we} reproduced this experiment and additionally compared to the average human rating that each 
past tense form received in A\&H's dataset. They found that the neural network from K\&C produced probabilities that were sensitive to random initialization -- showing high variance in the resulting correlations with humans -- and typically did not correlate better than the rule-based model from A\&H. They then designed an experiment where inflected forms were sampled from several different randomly initialized models, so that the frequencies of each form could be aggregated in a similar fashion to the adult production probabilities -- but the results still favored A\&H. They hypothesized that the model's overconfidence in the most likely inflection (i.e. the regular inflection class) leads to uncharacteristically low variance on predictions for unknown words.


\paragraph{German Noun Plural}
\citet[M\&al.]{mccurdy2020inflecting} applied an LSTM to the task of German noun plural inflection to investigate a hypothesis from \citet[M95]{marcus1995german}, who attributed the outputs of neural models to their susceptibility to the most frequent pattern observed during training, stressing that, as a result, neural approaches fail to learn patterns of infrequent groups.

German nouns inflect for the plural and singular distinction. There are five suffixes, none of which is considered a regular majority: /-(e)n/, /-e/, /-er/, /-s/, and /-$\emptyset$/.
 M95 had built a dataset of monosyllabic German noun wugs and investigated human behavior when inflecting the plural form, distinguishing between phonologically familiar environments (rhymes), and unfamiliar ones (non-rhymes). The German plural system, they argued, was an important test for neural networks since it presents multiple productive inflection rules, all of which are minority inflection classes by frequency. This is in contrast to the dichotomy of the regular and irregular English past tense. M\&al. collected their own human production probabilities and ratings for these wugs, and then compared those to LSTM productions. Humans were prompted with each wug with the neuter determiner to control for the fact that neural inflection models of German noun plurals are sensitive to grammatical gender \cite{goebel2000recurrent}, and because humans do not have a majority preference for monosyllabic, neuter nouns \cite{CLAHSEN1992225}.

The /-s/ inflection class, which is highly infrequent appears in a wide range of phonological contexts, which has lead some research to suggest it is the \textit{default} class for German noun plurals, and thus the regular inflection, despite its infrequent use. M\&al. found that it was preferred by humans in Non-Rhyme context more than Rhymes, but the LSTM model showed the opposite preference, undermining the hypothesis that LSTMs model human generalization behavior. /-s/ was additionally predicted less accurately on a held-out test set of real noun inflections when compared to other inflection classes.

They found that the most frequent inflection class in the training for the monosyllabic neuter contexts, /-e/, was over-generalized by the LSTM when compared to human productions. The most frequent class overall, /-(e)n/ (but infrequent in the neuter context), was applied by humans quite frequently to nonce nouns, but rarely by the LSTM. They additionally found that /-er/, which is as infrequent as /-s/, could be accurately predicted in the test set, and the null inflection /-$\emptyset$/, which is generally frequent, but extremely rare in the monosyllabic, neuter setting was never predicted for the wugs. We refer to \citet{mccurdy2020inflecting} for more details on the inflection classes and their frequencies, and additional discussion around their relevance to inflection behavior.

Ultimately, M\&al. reported no correlation with human production probabilities for any inflection class. They concluded that modern
neural networks still simply generalize the \textit{most frequent} patterns to unfamiliar inputs. 

\citet{dankers2021generalising} performed in-depth behavioral and structural analyses of German noun plural inflection by a \textit{unidirectional} LSTM \textit{without} attention. They argued that these modeling decisions made a more plausible model of human cognition. In a behavioral test they found that,  like humans but unlike M\&al., their model did predict /-s/ more for non-rhymes than for rhymes, but the result was not statistically significant. They also found that /-s/ was applied with a high frequency
and 
attributed this to sensitivity to word length. 
For a visual of 
all studies discussed in this section, see Figure \ref{fig:pst_tense_debate}.

\paragraph{Our Contribution}
Most work on modern neural networks discussed here analyzes the same bidirectional LSTM with attention and draws a mixture of conclusions based on 
differing experimental setups. \citet{dankers2021generalising} changed the LSTM-based architecture, and found somewhat different results for German number inflection, though they did not investigate correlations with human ratings nor production probabilities in the same way as previous work. The limited variation of architectures 
in previous studies as well as inconsistent methods of comparison with human behavior 
prevent us from drawing definite conclusions about the adequacy of neural networks as models of human inflection.

Here, we present results on a wider range of LSTMs and a Transformer \cite{vaswani2017attention} model for both English past tense and German number inflection. We ask which architecture is the best account for human inflection behavior
 and, following M\&al., investigate the actual model productions (and probabilities) for the German plural classes in order to qualitatively compare to human behavior. We additionally ask how architectural decisions for the LSTM encoder-decoder affect this correlation. Finally, we investigate the relationship between inflection \textit{accuracy} on the test set and \textit{correlation} with human wug ratings.

We find that the Transformer consistently correlates best with human ratings, producing probabilities that result in Spearman's $\rho$ in the range of 0.47-0.71 for several inflection classes, which is frequently higher than LSTMs. However, when looking closely at the Transformer productions, it displays behavior that deviates from humans similarly to the LSTM in M\&al., though to a lesser extent. While attention greatly increases LSTM \textit{accuracy} on inflection, we also find that it does not always lead to better correlations with human wug ratings, and that the directionality of the encoder has more complicated implications. Finally, we find that
there is no clear relationship between model \textit{accuracy} and \textit{correlation} with human ratings across all experiments, demonstrating that neural networks can solve the inflection task in its current setup without learning human-like distributions. While the Transformer experiment in this work demonstrates stronger correlations with human behavior, and \textit{some} more human-like behaviors than before, our findings continue to cast doubt on the cognitive plausibility of neural networks for inflection.


\section{Neural Morphological Inflection}

\subsection{Task Description}
The experiments in this paper are centered around a natural language processing (NLP) task called morphological inflection, which consists of generating an inflected form for a given lemma and set of morphological features indicating the target form.
It is typically cast as a character-level sequence-to-sequence task, where the characters of the lemma and the morphological features constitute the input, while the characters of the target inflected form are the output \cite{,kann-schutze-2016-single}:
\begin{align}
\textrm{PST ~ c ~ r ~ y} ~ \rightarrow ~ 
\textrm{c ~ r ~ i ~ e ~ d} \nonumber
\end{align}

Formally, let ${\cal S}$ be 
the paradigm slots expressed in a language and $l$ a lemma in the language. 
The set of all inflected forms -- or \textit{paradigm} -- $\pi$ of $l$ is then defined as:
\begin{equation}
  \pi(l) = \Big\{ \big( f_k[l], t_{k} \big) \Big\}_{k \in {\cal S}}
\end{equation}
$f_k[l]$ denotes the inflection of $l$ which expresses tag $t_{k}$, and $l$ and $f_k[l]$ represent strings consisting of letters from the language's alphabet $\Sigma$.

The task of morphological inflection can then formally be described as predicting the form $f_i[l]$ from the paradigm of $l$ corresponding to tag $t_i$.

\subsection{Models}

\paragraph{Rumelhart and McClelland}
The original model of \citet{rumelhart1985learning} preceded many of the features introduced by modern neural networks. For example, they use a feed-forward neural network to encode input sequences. This creates the requirement of coercing variable-length inputs into the fixed-size network. To solve this, they 
encode input words as fixed length vectors representing the phonological distinctive feature sets for each trigram in that word. The neural network is then trained to map the features of an input form to a feature vector of a hypothesized output form. The loss is computed between the input feature sets 
 and the the feature set for an inflected output form encoded in the same way.
At test time, they manually select candidate output forms for each input lemma in order to overcome the intractable decoding problem. The output form, then, is the candidate whose feature vector most closely resembles the model output.
Beyond decoding problems, the order of input characters is not encoded, and unique words are represented with potentially identical phonological features.

\paragraph{LSTM}
The LSTM architecture \cite{hochreiter1997long} overcomes several of the issues in \citet{rumelhart1985learning}, by way of a recurrent encoding and decoding mechanism, and reliance on character embeddings. 

We experiment with several variations of the LSTM encoder-decoder \cite{sutskever2014sequence,cho-etal-2014-learning} to test their behavior compared to humans. First, we vary \textit{directionality} of the encoder under the assumption that bidirectional encoding leads to higher accuracy, but a unidirectional encoder may better resemble human processing.
We additionally vary whether or not \textit{attention} is used. Attention is typically a crucial feature to attaining high inflection accuracy. We expect that the same may also be true for assigning a cognitively plausible probability to a nonce inflection, by supplying the model with a mechanism to focus on only the relevant parts of the inflection.


This yields 4 LSTM-based variations. We refer these models as \texttt{BiLSTMAttn} (BA; from K\&C, C\&al., and M\&al.), \texttt{UniLSTMAttn} (UA), \texttt{BiLSTMNoAttn} (BN), and \texttt{UniLSTMNoAttn} (UN; from \citet{dankers2021generalising}). 

\paragraph{Transformer} Finally, we present results for a Transformer sequence-to-sequence model \cite{vaswani2017attention}, following the implementation proposed for morphological inflection by \citet{wu-etal-2021-applying}. Unlike LSTM-based models, the transformer employs a \textit{self-attention} mechanism such that each character representation can be computed in parallel as a function of all other characters. The position of each character is encoded with a special positional embedding. This means that the relation between each character in a word can be represented directly, rather than through a chain of functions via the LSTM recurrence. It is considered to be state-of-the-art for morphological inflection in terms of accuracy, which makes it an important comparison for this study. Some work has called into question the cognitive plausibility of transformer self-attention in psycholinguistic experiments of word-level language models \cite{merkx2020human} -- claiming that the direct access it provides to past input is cognitively implausible. It is not clear though that these arguments apply to \textit{character-level} models for inflection, wherein words do not necessarily need to be processed one character at a time.

\paragraph{Hyperparameters}
We implement all LSTMs with pytorch \cite{NEURIPS2019_9015} and borrow hyperparameters from previous work on morphological inflection. For the LSTMs, we use the hyperparameters from K\&C, which were based on the tuning done by \citet{kann-schutze-2016-single}.
For the Transformer, we follow the hyperparameters from the best model in \citet{wu-etal-2021-applying}, but set label-smoothing to 0. In preliminary 
experiments, we found no significant impact of label smoothing on accuracy nor correlation with human behavior across inflection classes.

For all architectures, we follow C\&al. and train 10 randomly initialized models. At test time, we decode with beam search with a width of 12.
We train for up to 50 epochs because the architectures with fewer parameters tend to converge more slowly.

\paragraph{MGL}
A\&H implement the Minimal Generalization Learner (MGL), which learns explicit rules (e.g. insertion of /-\textipa{Id}/ if a verb ends in a /\textipa{t}/ or /\textipa{d}/) at varying levels of granularity. Each rule is associated with a confidence score for a given phonological environment based on its statistics in the train set. At test time, the rule with the highest confidence is applied to produce an inflection, and the confidences can be used to score various regular or irregular inflected forms. We compare to this model for English data, following previous work.

\begin{table*}[t] 
\small
\centering
\begin{tabular}{l|l|ll|ll|ll}
\toprule
 & Dev Acc & \multicolumn{2}{c}{Test Acc} & \multicolumn{2}{c}{Prod. Prob.} &  \multicolumn{2}{c}{Rating} \\ & & reg & irreg & reg & irreg & reg & irreg \\
\midrule
A\&H MGL & - & \textbf{99.7} & \textbf{38.0} & .33 & .30 & .50 & .49 \\ 
K\&C* & - & 98.9 & 28.6 & \textbf{.48} & .45 & - & - \\ 
C\&al. Agg.** & - & - & - & .45 & .19 & .43 & .31 \\ 
\texttt{BiLSTMAttn} & 93.33 & 97.48  (.65) & 9.05 (5.24) & .28 & .36 & .16 & .46 \\ 
\texttt{BiLSTMNoAttn} & 76.37& 82.72  (2.06) & 7.62 (3.33) & .14 & .44 & .23 & .35 \\ 
\texttt{UniLSTMAttn} & 92.45& 96.53  (.68) & 20.00 (4.38) & .35 & .41 & .40 & .32 \\ 
\texttt{UniLSTMNoAttn} & 73.49& 77.72  (1.64) & 10.48 (10.24) & .22 & .43 & .28 & .34 \\ 
\texttt{Transformer} & \textbf{94.88} & 99.21  (.53) & 10.95 (11.46) & .38 & \textbf{.47} & \textbf{.58} & \textbf{.58} \\ 
\bottomrule
\end{tabular}
\caption{\label{tab:en_results} English results for both regular (reg) and irregular (irreg) inflections for all architectures and metrics. Along with accuracy, we report Spearman's $\rho$ between average model rating and our two human metrics. Standard deviations are given in parentheses. \\
*Trained and tested a different random split,
**Trained and tested on all training data}
\end{table*}
\section{Experiments}
\subsection{Languages and Data}
We use the same data as previous work on English past tense, and German number inflection.

\paragraph{English}
We experiment with the English past tense data from A\&H, following both K\&C and C\&al. For training, we split the CELEX \cite{baayen1996celex} subset produced by A\&H, consisting of 4253 verbs (218 irregular), into an 80/10/10 random train/dev/test split following K\&C.\footnote{K\&C use a subset of A\&H, removing 164 regulars and 50 irregulars. We include them in our dataset.} We ensure that 10\% of the irregular verbs are in each of the development and test set. 

The English nonce words from A\&H, used for computing the correlation of model rating with human ratings and production probabilities, comprise 58 made-up verb stems, each of which has 1 regular and 1 irregular past tense inflection. 16 verbs have an additional irregular form (58 regulars and 74 irregulars total). 
All English data is in the phonetic transcription provided by A\&H.

\paragraph{German}
We also experiment with the German dataset from \citet{mccurdy2020inflecting}, who released train/dev/test splits consisting of 11,243 pairs of singular and plural nouns in the nominative case taken from UniMorph \cite{mccarthy-etal-2020-unimorph}. They added gender, the only inflection feature provided, by joining UniMorph with a Wiktionary scrape.

The German wugs come from M95, who built a set of 24 monosyllabic nonce nouns: 12 of which are rhymes -- resembling real words in their phonology, and 12 of which are non-rhymes -- representing atypical phonology.
Human ratings and production probabilities, however, are taken from M\&al., who administered an online survey to 150 native German speakers. Each participant was prompted with the nouns from M95 with the neuter determiner, and then asked to generate the plural form. Similar to A\&H, after producing a plural for each noun, participants were asked to rate the acceptability of each potential plural form on a 1-5 scale. In their analysis, M\&al. compare human and model behavior on 5 productive inflection classes, shown for our experiments in Table \ref{tab:de_results}. 

\subsection{Evaluation Metrics}
We evaluate models with respect to four metrics.

\paragraph{Accuracy} This refers to raw accuracy on a set of real inflections that the model has not seen during training. Crucially, only the top prediction of a given model 
is considered, 
and the model's probability distribution over all predictions does not affect the score. 

\paragraph{F1} We report F1 instead of accuracy for the German plural experiments following M\&al. Here we classify each inflected form with its suffix (e.g. /-s/), and classify inflections that do not conform to the 5 inflection classes from M\&al. as "other."

\paragraph{Production Probability Correlation} Like previous work \cite{kirov2018recurrent,corkery2019we,mccurdy2020inflecting}, we compare model output probabilities with production probabilities from humans. The production probability of a form is calculated by counting all forms produced for a given lemma, and then normalizing them to obtain a probability distribution of the human productions. In keeping with most previous work and because we do not expect a linear relationship with the model ratings, we report Spearman's $\rho$. This is calculated \textit{within} each inflection class, meaning that, e.g., for English we report a regular and an irregular $\rho$. For example, the regular $\rho$ for the set of lemmas \{rife, drize, flidge\} would be computed from the vector containing probabilities of the forms \{rifed, drized, flidged\} under the model, against the corresponding vector with human probabilities.

\paragraph{Rating Correlation} Finally, we compare model ratings to the average human rating of each form, again reporting $\rho$ within inflection class. Here, rather than normalizing over production frequencies, humans were prompted with an inflection for a given lemma and asked to rate it on a scale that differed slightly between datasets. For each lemma, we thus get an average probability for a regular form, as well as for an irregular form.

\subsection{Neural Network Wug Test}\label{sec:NN-wug-test}
In order to compare 
our models to humans, 
we compute analogous values to the human ratings and production probabilities. We investigate two strategies: normalizing the inflected form counts output by our models, and computing the average probability of each form under our models.

\paragraph{Model Production Probability}
Previous work \cite{corkery2019we,mccurdy2020inflecting} decoded outputs from multiple models and aggregated the resulting forms: given a lemma and a set of $n$ models trained with different random seeds, an inflected form is sampled from each model, resulting in forms ${f_1, ..., f_n}$, where forms need not be unique.
The frequency of each form is then normalized to obtain a probability distribution. For example, given the nonce lemma \textit{rife}, the probability of the past tense form \textit{rifed} is computed as

\begin{equation*}
   \frac{1}{n} \sum_{i=1}^{n} \begin{cases}
    1,& \text{if } f_i = \text{rifed}\\
    0,              & \text{otherwise}
\end{cases}
\end{equation*}

C\&al. propose a version of this in their aggregate model, in which they sample 100 forms from each model
, and normalize the resulting form frequencies. M\&al., who instead train 25 randomly initialized models, perform the same aggregation over the \textit{top} prediction of each model. We take the approach of M\&al. (though we train only 10 models) to investigate model productions qualitatively. This metric is intuitively similar to quantifying human production probabilities if we consider one model to be one human participant.

\begin{table*}[t] 
\small
\centering
\begin{tabular}{l|c|llllll}
\toprule
 & Dev Acc. &  \multicolumn{6}{c}{Test F1} \\ & & /-(e)n/ & /-e/ & /-$\emptyset$/ & /-er/ & /-s/ & other \\
\midrule
M\&al. & 92.10 & \textbf{95.00}  & 87.00 & 92.00  & \textbf{84.00}  & \textbf{60.00} & 42.00  \\
\texttt{BiLSTMAttn} & 89.37 & 93.93  (0.6)  & \textbf{88.08  (0.9)}  & 92.43  (0.6)  & 79.07  (5.1)  & 51.75  (4.6)  & 45.36  (4.0) \\
\texttt{BiLSTMNoAttn} & 54.65 & 74.16  (1.9)  & 63.56  (2.4)  & 75.57  (2.1)  & 51.26  (3.7)  & 29.58  (7.4)  & 9.07  (0.6)  \\
\texttt{UniLSTMAttn} & 86.40 & 93.39  (0.6)  & 87.35  (1.0)  & 92.49  (1.1)  & 69.78  (5.3)  & 52.36  (4.5)  & 44.06  (5.8) \\
\texttt{UniLSTMNoAttn} & 48.71 & 69.69  (2.2)  & 58.31  (2.4)  & 71.98  (1.7)  & 46.64  (5.2)  & 32.54  (7.7)  & 8.08  (0.4) \\
\texttt{Transformer} & 91.04 & 92.93  (0.4)  & 87.81  (0.7)  & \textbf{93.86  (0.3)}  & 65.44  (4.7)  & 57.89  (2.0)  & \textbf{57.47  (4.5)} \\
\bottomrule
\end{tabular}
\caption{\label{tab:de_f1s} Average German F1s on all German plural inflections for all architectures. Standard deviation is given in parentheses. Dev accuracy for our experiments were decoded greedily.}
\end{table*}

\begin{table*}[t] 
\centering
\small
\begin{tabular}{l|cccccc|cccccc}
\toprule
 & \multicolumn{6}{c}{Prod. Prob} &  \multicolumn{6}{c}{Rating} \\ & /-(e)n/ & /-e/ & /-$\emptyset$/ & /-er/ & /-s/ & avg. & /-(e)n/ & /-e/ & /-$\emptyset$/ & /-er/ & /-s/ & avg. \\
\midrule
M\&al. & .28  & .13  & -  & .05  & .33 & .20 & -  & - & - & - & - & -\\
\texttt{BiLSTMAttn} & .11  & .08  & -.14  & .24  & .38 & .20 & .36  & .44  & .06  & .36  & .39 & .32 \\
\texttt{BiLSTMNoAttn} & \textbf{.44}  & .08  & -.12  & .27  & .39 & \textbf{.30} & \textbf{.51}  & .16  & -.29  & .30  & .31 & .20 \\
\texttt{UniLSTMAttn} & .09  & .16  & -.13  & \textbf{.36}  & .39 & .25 & .22  & .27  & -.16  & .46  & .44 & .25 \\
\texttt{UniLSTMNoAttn} & .14  & .15  & \textbf{.08}  & .17  & .23 & .15 & .24  & .16  & -.17  & .05  & .20 & .10 \\
\texttt{Transformer} & .11  & \textbf{.30}  & -.13  & .28  & \textbf{.50} & .20 & .48  & \textbf{.59}  & \textbf{.15}  & \textbf{.50}  & \textbf{.71} & \textbf{.49} \\
\bottomrule
\end{tabular}
\caption{\label{tab:de_results} German wugs Spearman's $\rho$ for the average rating of each model with human production probabilities (left) and average human ratings (right). We report the macro average (avg.) over all inflection classes 
for both.}
\end{table*}
\paragraph{Model Rating}
Because the aggregate outputs method considers only the \textit{most likely} prediction aggregated over the same architecture trained on the same dataset, we expect the prediction to typically be the same for each model. We instead report correlations with the \textit{probability} of inflected forms under each model in Tables \ref{tab:en_results} and \ref{tab:de_results}.
K\&C correlate this value with human production probabilities, and C\&al. use this method in an experiment to compute individual model ratings.

More formally, given a lemma $l$ and an inflected form $f$ of length $k$, we compute 

\begin{align}
    p(f \mid l) &= p(f_{1}, ..., f_{k} \mid l) \\
    &= \prod_{1}^{k} p(f_{i} | f_{i-1}, l)
    \label{eq:form_rating}
\end{align}

Where $f_{i}$ is the $i$th character of $f$. We force the model to output each inflected form $f$ to get its probability. In practice, we modify Equation \ref{eq:form_rating} to compute a length-normalized probability because $p(f \mid l)$ becomes smaller as $f$ increases in length. For $f$ of length $k$, we have

\begin{equation}
    p(f \mid l) = \sqrt[k]{\prod p(f_{i} | f_{i-1}, l)}
    \label{eq:normalized_form_rating}
\end{equation}

\begin{figure*}[t]
    \small
    \centering
    \includegraphics[width=0.85\linewidth]{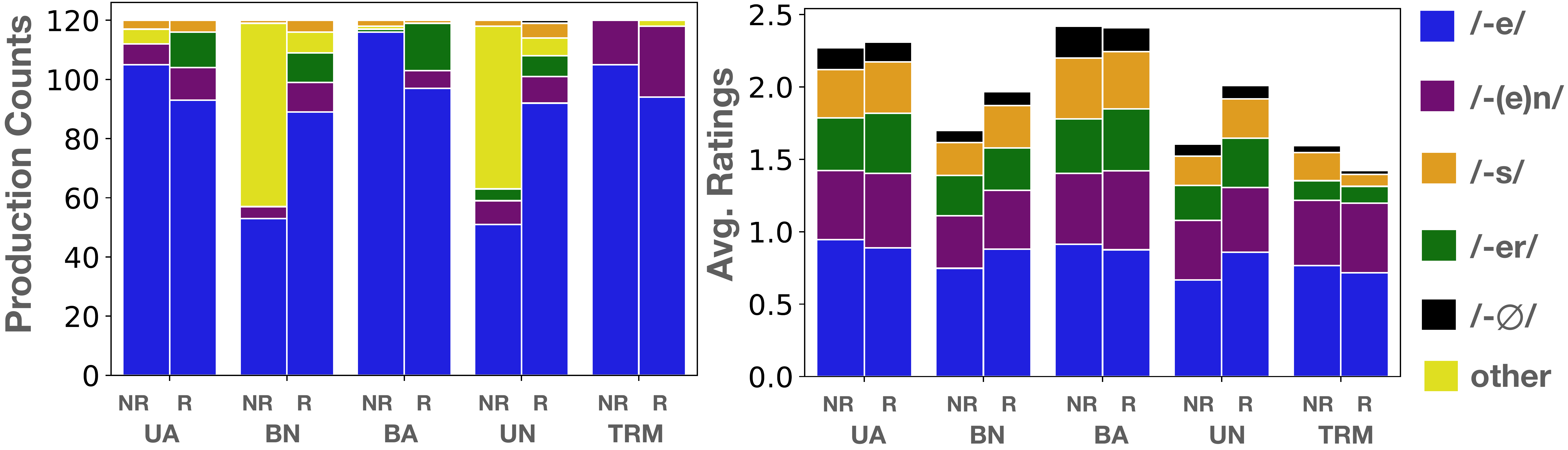}
    \caption{German plural productions (left) and average probabilities (right) for each architecture in Rhyme (R) and Non-Rhyme (NR) contexts for all lemmas and all random initializations. Shorthands are used for architectures -- UA refers to \texttt{UniLSTMAttn}, whereas BN refers to \texttt{BiLSTMAttn}, for example.}
    \label{fig:M95_prods}
\end{figure*}

We expect computing ratings in this way to be similar to the aggregate model of C\&al. described above. That is, the probability of a form $f$ computed by aggregating $n$ forms from a single model's probability distribution should approach $p(f \mid l)$, as $n \rightarrow \infty$. Finally, we compute the average probability of a form from all 10 randomly initialized models, and refer to it as the model rating.

\section{Results}
We present experimental results in Tables \ref{tab:en_results}, \ref{tab:de_f1s}, and \ref{tab:de_results} in terms of both inflection accuracy,
and correlation with human behavior -- our main focus. All correlations for neural models trained in this work are given with respect to \textbf{model rating}, and not the model production probability. We report results from training MGL on our data, and include the results reported K\&C, C\&al., and M\&al. in the appropriate tables for reference.
\subsection{English}
For English, many of our models correlate better for irregulars than regulars, unlike previous work for which the strongest correlations occurred for regular verbs. As we do not have the same train and test splits, it is difficult to draw conclusions from this result. We predominately focus on performance differences between the models trained in our experiments including MGL.

\paragraph{Accuracy}
The accuracies from this experiment generally reflect our expectations from prior work. The Transformer attains the highest test accuracy, LSTMs with attention always achieve higher accuracy than those without, and
bidirectional LSTMs show modest improvements over their unidirectional counterparts.
However, the unidirectional LSTMs outperform bidirectional counterparts for irregular accuracy (+2.86 and +10.95).
Additionally, the Transformer has a low irregular accuracy, though with a very high standard deviation over all 10 runs, indicating at least one run was an outlier with much higher accuracy.

\paragraph{Correlation}
The trend in accuracy for attentional LSTMs is not strictly true for correlation. LSTMs \textit{without} attention typically correlate with humans slightly better than their attentional counterparts for irregulars. Additionally, unidirectional models result in higher \textit{regular correlations}, which is in contrast to the higher irregular accuracy. Irregular correlations are fairly similar across LSTMs with the exception of the \texttt{BiLSTMAttn} correlation with human ratings, which is much higher than the other LSTM correlations. We also reproduce previous results showing that A\&H's rule-based model, MGL, is better correlated than any LSTM model. The transformer, however, is correlated most highly with humans among all experiments that we ran. 

\subsection{German}
We refer to F1 in Table \ref{tab:de_f1s}, and correlation with humans in Table \ref{tab:de_results}. Notably all models typically correlate better with human ratings than with production probabilities, though those two metrics have a positive linear relationship ($r$=0.75). Intuitively, the task of assigning a probability to a form is more like the human rating task than decoding the single most likely form. We present a graph of model production probabilities and model ratings for the German wugs in Figure \ref{fig:M95_prods}.

\paragraph{F1}
F1 scores follow a similar trend to English. In contrast to the very small performance gap in \citet{dankers2021generalising}, LSTMs with attention clearly perform better in terms of F1 than without -- though our training dataset from M\&al. is much smaller than the one they used, which might amplify the gap. Directionality has much less effect on F1 than attention for German, with the unidirectional LSTMs actually outperforming the bidirectional ones for the infrequent /-s/ class in our experiments. The Transformer attains high (though not necessarily the highest) F1 scores for every class.

\paragraph{Correlation}
The Transformer clearly correlates most highly with human ratings, attaining a moderate correlation (0.48-0.59) for /-e/, /-(e)n/, and /-er/, and a high correlation (0.71) for /-s/. All architectures correlate poorly with /-$\emptyset$/, despite very high F1. Looking more closely at /-$\emptyset$/, it consistently receives very low ratings (as is the case for human ratings), and it was never produced as a model's best output as can be seen in Figure \ref{fig:M95_prods}. However, there are only 3 /-$\emptyset$/ inflections in the training data that fit the same phonological context as the wugs. Across all contexts though, /-$\emptyset$/ is a very common inflection in the training data, which explains its high accuracy on the test set.

There is no clear trend between LSTMs in terms of correlation with human production probability, with rather low $\rho$ overall. However in the case of human ratings, LSTMs with attention always correlate better than those without, with the exception of the most frequent class overall in the training data, /-(e)n/.  \texttt{BiLSTMNoAttn} is most strongly correlated for /-(e)n/, in contrast to its lower F1 -- demonstrating that removing attention leads to a lower F1, but also to a more human-like probability of /-(e)n/.

Regarding directionality, unidirectional LSTMs always outperform their bidirectional counterparts for the infrequent /-s/ class in our experiments. \texttt{UniLSTMAttn} correlates better with humans than any LSTM for the infrequent classes /-er/ and /-s/. However, \texttt{BiLSTMAttn} has the highest correlation for the frequent /-e/ and /-(e)n/.


\section{Analysis}
We mainly analyze the correlation between (average) model ratings and human \textit{ratings}. 
\textit{We find that the Transformer correlates best with human ratings with few exceptions}, indeed it attains a statistically significant positive correlation for all inflection classes in both languages, with the exception of /-$\emptyset$/ in German. It is also highly accurate, as in previous work \cite{wu-etal-2021-applying}.

Regarding LSTM architectural decisions, unsurprisingly, attention and bidirectionality typically increase accuracy in both languages. The positive effect of attention is similar for correlations with some exceptions. Attention almost always leads to better correlations in German, with the interesting exception of /-(e)n/. Given that humans rate /-(e)n/ most highly on average, the higher correlation could be because without attention, LSTMs are very sensitive to the high /-en/ frequency in the training set. The attentional LSTMs might learn the monosyllabic, neuter context that applies to the wugs, for which there are very few /-(e)n/ training examples.
Despite slightly higher accuracy for bidirectional LSTMs, unidirectional LSTMs tend to attain higher correlations with both human metrics for English, especially for the more frequent \textit{regular} inflections.

Conversely in German, the bidirectional LSTMs correlate better for the more frequent /-(e)n/ and /-e/ classes, but \texttt{UniLstmAttn} correlates better for the rarer /-er/ and /-s/ classes. 
The dichotomy between just one highly productive class in English and several productive classes in German may explain the first observation: if unidirectional LSTMs \textit{overfit} to the frequent class, then they might appear to correlate better in English, but not German. However, this would not explain the German class correlations for \textit{infrequent} inflections, which could be explored in future work.

\paragraph{German Model Productions}
The model production counts in Rhyme versus Non-Rhyme contexts were important for the conclusion in M\&al. that \texttt{BiLSTMAttn} is not a good model of human behavior. We thus investigate this in Figure \ref{fig:M95_prods}.

Most of the criticisms from M\&al. apply to the productions in our experiments as well. One new observation is that, without attention, LSTMs predict many "other" forms for NR contexts, but not for R. This likely means that Non-Rhymes lead to decoding errors for these models due to the unfamiliar context. Additionally, despite several behaviors that differ from humans in the Transformer productions, its second most produced inflection class is /-(e)n/, like humans, and unlike any LSTM model.
The right side of Figure \ref{fig:M95_prods} instead displays the average model rating of each inflection class, on which we base our correlations in Tables \ref{tab:en_results} and \ref{tab:de_results}.

The average model rating of an inflection class represents the probability assigned to it averaged over all 10 randomly initialized models and all 24 lemmas.
The /-e/ inflection accounts for a much smaller amount of the probability mass on average than its production probability.
The preference for /-e/ in the NR context, which diverges from human ratings, is smaller by this metric for the Transformer and LSTMs with attention. Furthermore, /-(e)n/ has a more reasonable average probability for most models when compared to the human ratings in M\&al., despite the preference for Rhymes, which diverges from human behavior. However, for /-s/ the Transformer shows a much higher average probability for Non-Rhymes than for Rhymes, which is more in line with human ratings.

Overall, this means model \textit{ratings} of German noun plurals look more similar to human ratings than model \textit{productions} do to human productions. The Transformer is a better account for human behavior than the LSTM, though it still diverges in some ways. \citet{dankers2021generalising} warned that the /-s/ behavior may be explainable by a simple heuristic though, so this behavior may not actually indicate cognitive plausibility.
\begin{table}[t] 
\small
\centering
\begin{tabular}{l|lllllll}
\toprule
& reg & irreg & /-(e)n/ & /-e/ & /-$\emptyset$/ & /-er/ & /-s/ \\
\midrule
$r$ &  0.44 & -0.31 & 0.01 & 0.80 & 0.73 & 0.70 & 0.83 \\
\bottomrule
\end{tabular}
\caption{\label{tab:acc_rating_corrs_by_class} Pearson $r$ between model acc. (or F1), and correlation with human ratings within infl. class (n=5).}
\end{table}
\begin{table}[t] 
\small
\centering
\begin{tabular}{l|llllll}
\toprule
& \texttt{BA} & \texttt{BN} & \texttt{UA} & \texttt{UN} & \texttt{Trm} \\
\midrule
$r$ & -0.57 & -0.33 & -0.37 & -0.39 & -0.38 \\
\bottomrule
\end{tabular}
\caption{\label{tab:acc_rating_corrs_by_model} Pearson $r$ between model acc. (or F1), and correlation with human ratings within model (n=7).}
\end{table}
\paragraph{Accuracy vs. Correlation}
The task of predicting the \textit{most likely} inflection for an unknown word (measured by accuracy or F1) is not the same as rating \textit{multiple inflections} (measured by Spearman's $\rho$). We thus investigate the relationship between these two tasks by measuring Pearson's $r$ between them to see if better inflection models in terms of accuracy are also more human-like. First, we consider the relationship for all models and inflection classes in both datasets and find no correlation ($r$ = -0.17, n=35). However, some inflection classes or models may behave differently than others. We refer to Table \ref{tab:acc_rating_corrs_by_model} to investigate this relationship within each architecture. In Table \ref{tab:acc_rating_corrs_by_class}, we check the correlation within each inflection class. There is not sufficient data to draw statistically significant conclusions in either case, but the correlations that we report can still characterize the relationship in our experiments. We find that all architectures show a negative correlation.
This implies that models are more accurate for inflection classes on which they correlate poorly with humans, and vise versa. However, Table \ref{tab:acc_rating_corrs_by_class} shows that all German inflection classes have a positive correlation between the two metrics, with the exception of /-(e)n/. This is likely because /-e(n)/ is highly frequent in the training set, but is less suitable for the monosyllabic, neuter wugs. Neither English inflection class shows a strong relationship, though. 

\section{Conclusion}
We ask which neural architecture most resembles human behavior in a wug test.
We introduce results on a wider range of architectures than previous work and find that the Transformer, a state-of-the-art model for morphological inflection, frequently 
correlates best with human wug ratings. 
Despite this, a closer look at model ratings and productions on German plural inflection shows that 
neither model closely resembles human behavior. We also find that, while attention is crucial for LSTM inflection accuracy, it does not always lead to higher correlations with humans. Additionally, the often less accurate unidirectional model sometimes correlates better than its bidirectional counterpart, especially in the case of infrequent German plural classes. Finally, while for some 
inflection classes more accurate models correlate better with humans, there is no clear relationship between the two metrics overall. Future work might consider behavior when hyperparameters are tuned to maximize plausibility of the probability distribution rather than accuracy.
Additionally, these results motivate a closer look at the effect of LSTM encoder directionality with respect to inflection class frequency.

\section*{Limitations}

This work is limited by the scope of languages and inflection categories that our models are tested on. We present results for two specific inflection categories in two languages. Previously, \citet{mccurdy-etal-2020-conditioning} ran experiments on neural network behavior for the German plural wugs used here, which brought into question some of the conclusions found in prior work for English past tense inflection. We thus believe that expanding this work to new inflection phenomenon and new languages may introduce results where the findings here do not necessarily hold.

\section*{Acknowledgments}
We would like to thank Kate McCurdy and Yohei Oseki for their input to and feedback on early stages of this work. We would also like to thank the anonymous reviewers,  Abteen Ebrahimi, and Ananya Ganesh for their feedback on drafts of this paper. This research was supported by the NSF National AI Institute for Student-AI Teaming (iSAT) under grant DRL 2019805. The opinions expressed are those of the authors, and do not represent views of the NSF.

\bibliography{anthology,custom}
\bibliographystyle{acl_natbib}


\clearpage
\appendix

\label{sec:appendix}
\counterwithin{figure}{section}

\section{Individual Model Variance}
\begin{figure*}[h!]
\small
\centering
     \begin{subfigure}[t]{.45\textwidth}
         \centering
         \includegraphics[width=\textwidth]{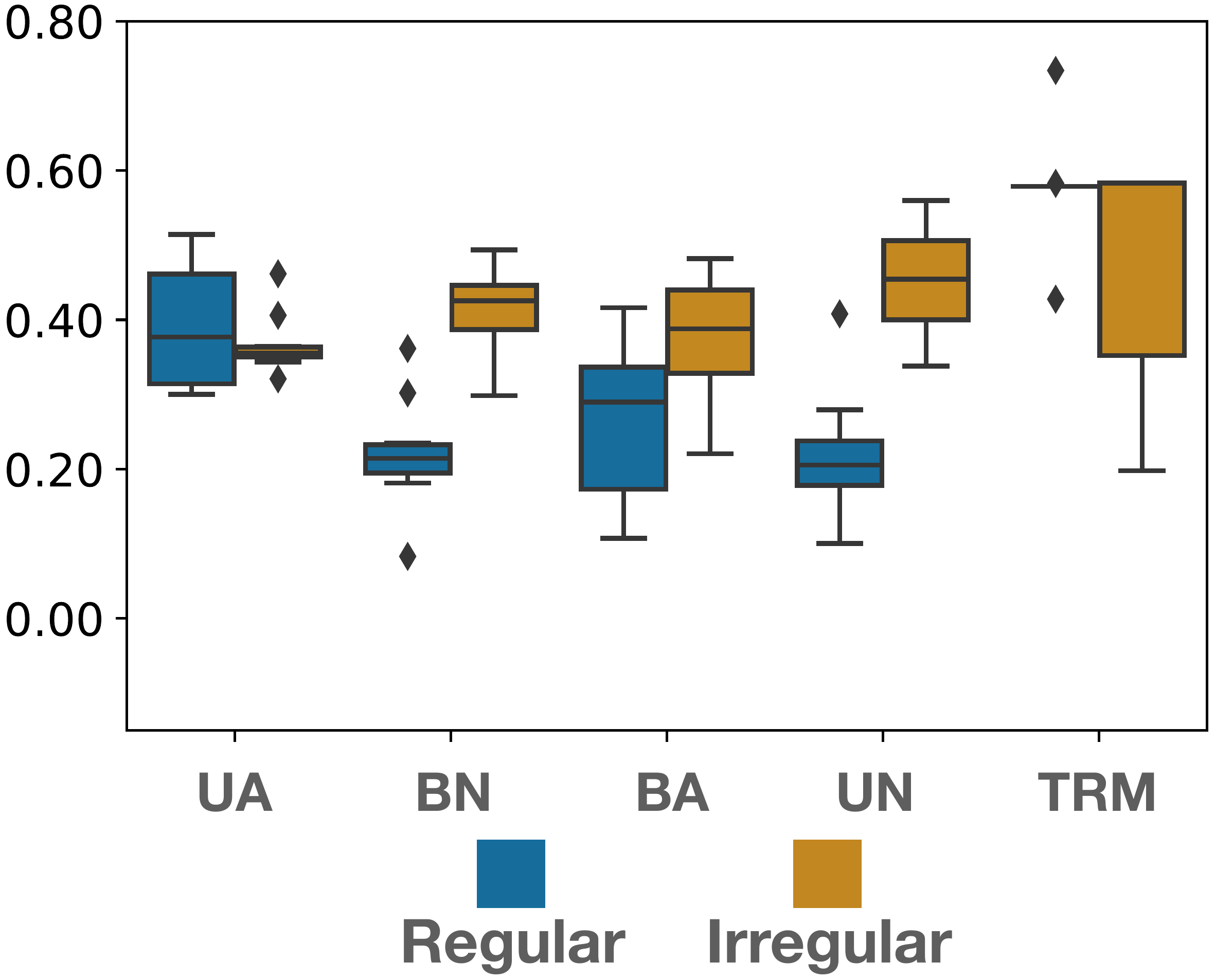}
         \caption{English past tense}
         \label{fig:AH_variance}
     \end{subfigure}
     \hfill
     \begin{subfigure}[t]{.45\textwidth}
         \centering
         \includegraphics[width=\textwidth]{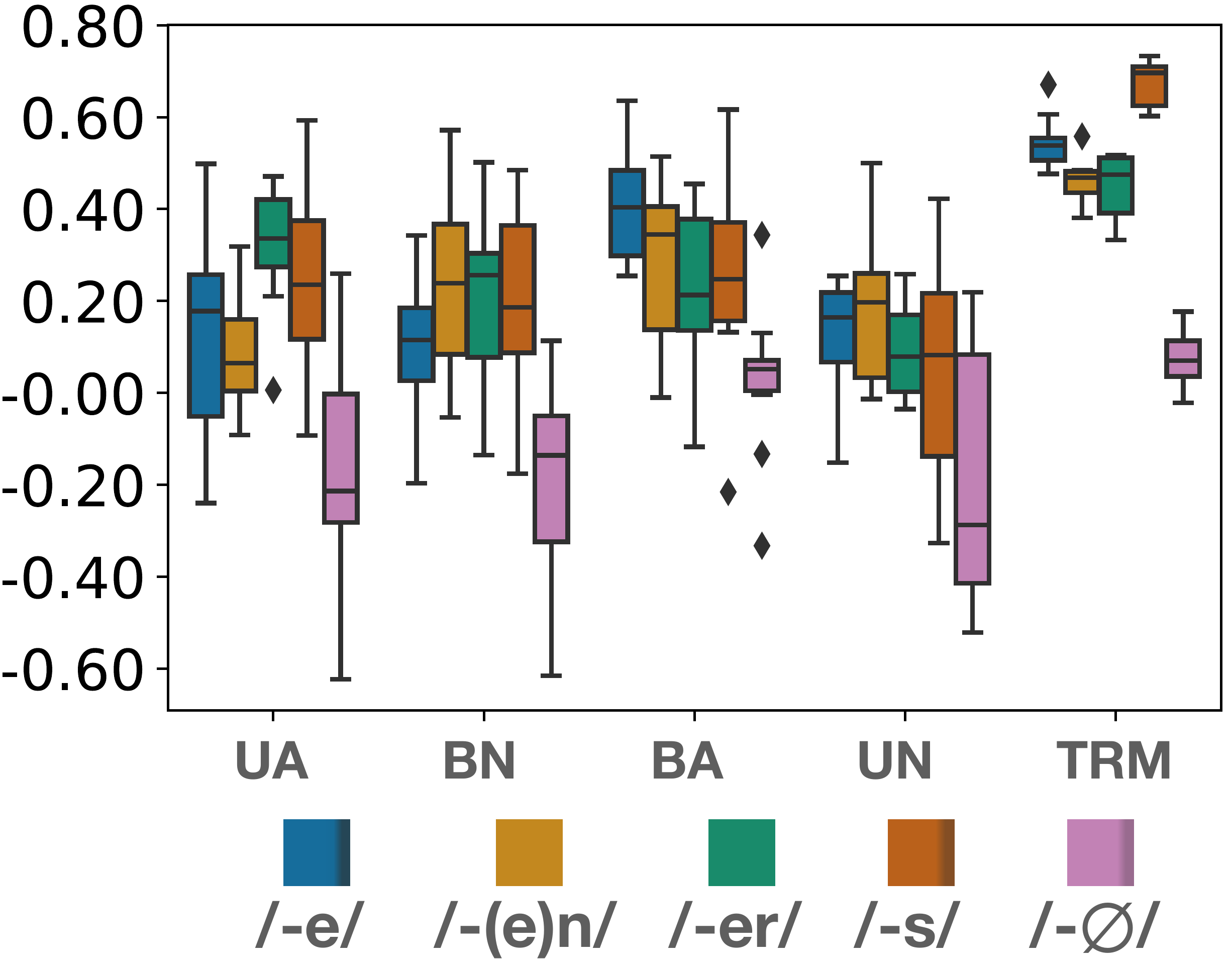}
         \caption{German plural}
         \label{fig:M95_variance}
     \end{subfigure}
     \caption{Boxplots of Spearman's correlation for individual models with respect to average human ratings}
     \label{fig:variances}
\end{figure*}

In figure \ref{fig:variances}, we show the variance, via boxplots, when correlating with human \textit{ratings}. Models typically have higher correlations with ratings than with production probabilities, but the two are linearly related in our results.  Similar to the findings of C\&al., who compared to production probabilities, we find that individual \texttt{BiLSTMAttn} models vary quite a bit with respect to correlation with humans. For English, some models vary far less, for example \texttt{BiLSTMAttn} has a much lower variance with respect to both regulars and irregulars than \texttt{BiLSTMAttn}. Similarly, the Transformer often correlates the same across different random initializations, with the exception of a few outliers. Turning to the German boxplots in \ref{fig:M95_variance}, we see similarly low variance for the transformers, and typically higher variance for most LSTMs. For architectures that vary more, i.e. LSTMs, we often see a higher correlation when the ratings are first averaged (as reported in Table \ref{tab:en_results} and \ref{tab:de_results}), but the same is often not true for English. 

\begin{table}[t] 
\small
\centering
\begin{tabular}{l|l}
\toprule
Model & Hyperparams. \\
\midrule
\texttt{BiLSTMAttn} & 0.93M \\
\texttt{BiLSTMNoAttn} & 0.90M \\
\texttt{UniLSTMAttn} & 0.56M \\
\texttt{UniLSTMNoAttn} & 0.54M \\
Transformer & 7.41M \\
\bottomrule
\end{tabular}
\caption{\label{tab:model_params} Number of parameters in each model.}
\end{table}

\onecolumn
\begin{figure*}[h!]
    \centering
    \includegraphics[width=0.8\textwidth]{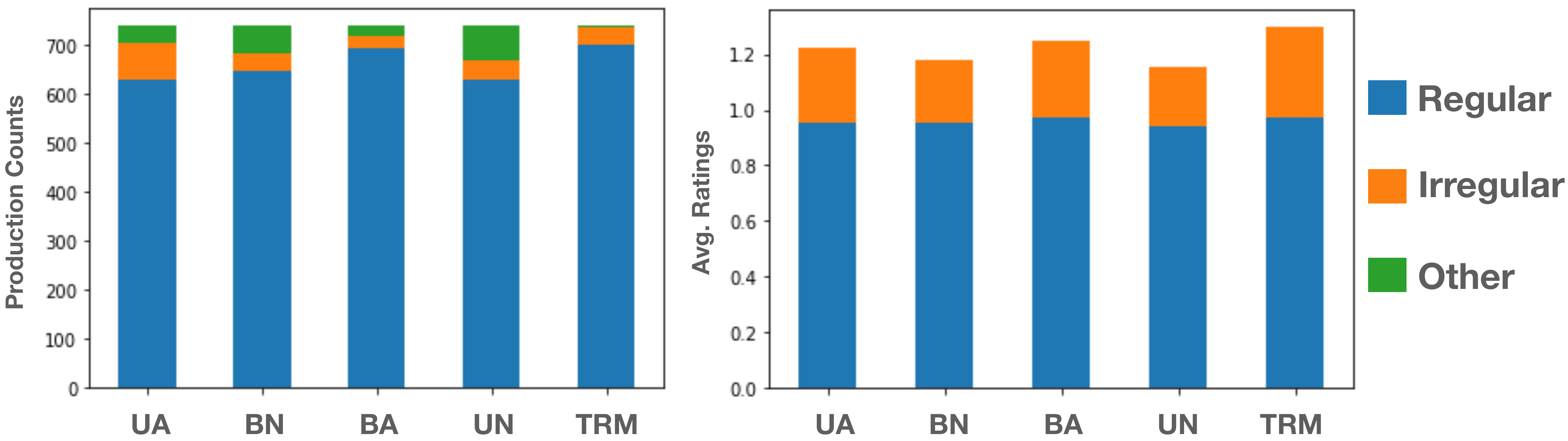}
    \caption{English past tense productions (left) and average probability (right) for each architecture for all lemmas and all random initializations.}
    \label{fig:AH_productions}
\end{figure*}

\end{document}